# AttentionSmithy: A Modular Framework for Rapid Transformer Development and Customization

Caleb Cranney, Jesse G. Meyer

Smidt Heart Institute, Department of Computational Biomedicine, Advanced Clinical Biosystems Research Institute, and Board of Governors Innovation Center

Cedars Sinai Medical Center, Los Angeles CA, 90048

Correspondence to caleb.cranney.app@gmail.com or jessegmeyer@gmail.com

**Abstract**
Transformer architectures have revolutionized a broad spectrum of AI applications by leveraging attention mechanisms for parallelized and long-range sequence processing. Despite their remarkable success, building and customizing transformers remains prohibitively complex for many domain experts who lack deep knowledge of low-level implementations. We introduce AttentionSmithy, a modular software package that lowers the barrier to transformer innovation by decomposing key components—attention modules, feed-forward networks, normalization layers, and positional encodings—into reusable building blocks. By disentangling architectural elements into well-defined interfaces, users can rapidly prototype, adapt, and evaluate transformer variants without extensive coding overhead. Our framework supports four distinct positional encoding strategies (sinusoidal, learned, rotary, and ALiBi) and integrates seamlessly with neural architecture search (NAS) for automated design exploration. We validate AttentionSmithy by replicating the original "Attention Is All You Need" transformer under resource constraints, demonstrating near state-of-the-art performance on a machine translation task. Leveraging the package's integrated NAS capability, we made the unexpected discovery that machine translation performance is maximized by combining all available positional encoding methods—highlighting the complementary benefits of each strategy. We further illustrate AttentionSmithy's adaptability through gene-specific modeling, where a variant of a BERT-style architecture achieves over 95% accuracy on downstream cell type classification tasks using ranked transcriptomic data. These case studies underscore AttentionSmithy's core advantage: enabling specialized experimentation across diverse application domains—from natural language processing to genomic analysis—by obviating the need for labor-intensive, low-level framework manipulation. We anticipate that AttentionSmithy will serve as a foundation for creative transformer-based solutions, expediting research and development in numerous scientific and industrial fields.

# Introduction

The transformer architecture [1] has revolutionized artificial intelligence, fundamentally changing how we approach sequence processing tasks across diverse domains. As transformer-based models continue to drive technological advancement and reshape societal interactions [2], there is growing interest in adapting these architectures for specialized applications. However, customizing transformer architectures remains a significant challenge, requiring deep expertise in both the theoretical foundations and implementation details. This complexity creates a barrier for domain experts who could otherwise leverage transformer capabilities for novel applications.

## Transformer Architecture Fundamentals

While traditional recurrent neural networks like Long Short-Term Memory (LSTM) networks [3] excelled at processing sequential data, they faced inherent limitations in parallelization and capturing long-range dependencies. The transformer architecture introduced by Vaswani et al. [1] overcame these constraints through its innovative attention mechanism, enabling unprecedented advances in natural language processing [4], computer vision [5], healthcare applications [6], molecular science research [7], and genomic analysis [8].

The basic building blocks of a transformer include [**Figure 1A**]:

1. Multi-head attention layers that compute and weigh relationships between all sequence elements in parallel
2. Feed-forward neural networks that process these relationships through non-linear transformations
3. Layer normalization components that stabilize training by normalizing activations across features
4. Residual connections that facilitate gradient flow and help preserve and reuse features from earlier layers
5. Positional encoding mechanisms that preserve sequence order information by encoding relative or absolute positions

## Positional Encoding Strategies

A crucial aspect of transformer architectures is their handling of sequential information through positional encodings. Without such encodings, transformers would treat input sequences as unordered sets of tokens, losing critical information about both absolute positions (where exactly a token appears in the sequence) and relative positions (how tokens are ordered with respect to each other). For instance, the sentences "the dog chased the cat" and "the cat chased the dog" contain identical tokens but convey opposite meanings, while "chased cat dog the the" is syntactically invalid – distinctions a transformer could not make without position information. While the self-attention mechanism excels at capturing relationships between tokens, it is

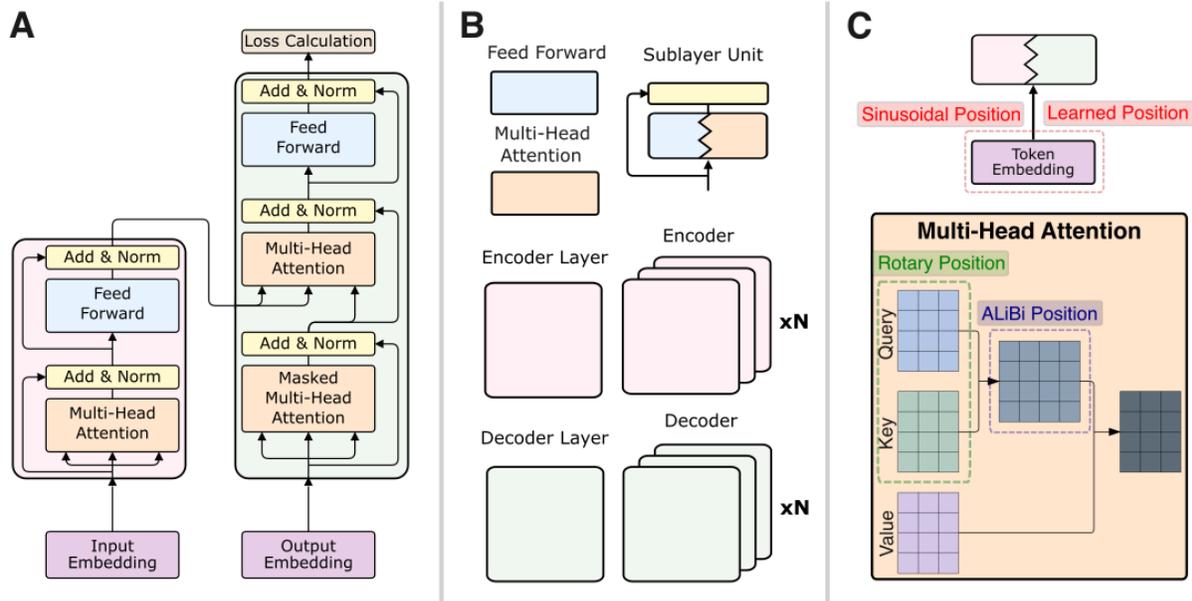

**Figure 1. Components of the transformer model architecture as coded in AttentionSmithy**. (A) The original transformer architecture introduced by Vaswanii et al. [1] (B) Labelled AttentionSmithy classes include implementations of the feed forward network, multi-head attention, sublayer unit (consisting of layer normalization and a residual connection surrounding an exchangeable feed forward network or multi-head attention class), encoder/decoder layers, and full encoders/decoders (which consist of an "N" number of layers determined by the end user). (C) Explicitly encoding position is a requirement of transformer models, but how position is encoded varies dramatically in implementation requirements from strategy to strategy. Four common strategies and their implementation details are outlined. The sinusoidal and learned positional embedding methods (red) involve directly adding vectors representing absolute positions to token embeddings before entering encoder or decoder layers. The rotary positional embedding method (green) requires adjusting the query and key matrices of the attention calculation directly. The ALiBi position embedding method (blue) adds negative values to the output of the query and key matrix multiplication that accumulate over greater positional distances.

inherently permutation-invariant, necessitating an explicit method to encode positional context. Several strategies have emerged, each with unique implementation requirements:

Sinusoidal positional encodings [1] and learned positional embeddings [9] operate by adding position-specific vectors directly to input token embeddings. This straightforward approach allows for easy implementation but may have limitations in capturing relative positions effectively.

Rotary positional embeddings [10] take a different approach, modifying the attention computation itself by applying rotation transformations to the query and key matrices. This method has shown particular promise in capturing relative positioning information while maintaining consistent attention patterns across sequence lengths.

ALiBi positional encodings [11] introduce position-specific bias terms to the attention score matrix, effectively modulating the attention weights based on relative positions. This approach has demonstrated advantages in extrapolating to longer sequences than those seen during training.

## Architectural Experimentation and Search

The modular nature of transformer architectures presents significant opportunities for systematic architectural exploration. Neural architecture search (NAS) has emerged as a promising approach for discovering optimal neural network configurations, but its application to transformers remains limited. While specialized NAS frameworks have been developed for transformers [12], [13], they are typically purpose-built for specific research objectives, making it difficult for practitioners to adapt them for novel transformer architectures and unique application domains. Traditional implementations often tightly couple architectural elements, making it challenging to define a comprehensive search space that includes variations in attention mechanisms, positional encodings, and feed-forward networks.

## Current Tools and Limitations

While popular libraries like HuggingFace [14], PyTorch [15], [16], and TensorFlow [17], [18], [19] provide implementations of standard transformer variants, they typically offer limited flexibility for architectural customization. These frameworks often implement positional encoding methods as integral parts of specific architectures, making it difficult to experiment with different encoding strategies or architectural variations. This tight coupling not only impedes manual experimentation but also makes it challenging to implement automated architecture search strategies. The complexity of transformer customization calls for a modular, component-based approach aligned with established software design principles [20], [21].

We present a novel software package called AttentionSmithy aimed at democratizing transformer development for domain experts. Drawing inspiration from design patterns [22], AttentionSmithy implements a "building block" architecture that promotes code reuse and maintainability while preserving architectural clarity. This modular design simultaneously serves multiple purposes: it facilitates rapid prototyping, enables systematic architecture search, and serves as an educational tool for understanding transformer architectures. By abstracting transformer components into reusable modules, we enable rapid experimentation while maintaining architectural clarity. Furthermore, AttentionSmithy includes positional encoding strategy implementations that empower developers to select the methods most suitable for their use case. Our approach addresses a critical gap in the current landscape, where despite the proliferation of transformer variants [23], implementing customized architectures remains a complex undertaking that often requires building from scratch.

# Methods

## Software Architecture

Our software package implements a component-based design philosophy to facilitate the creation of customized transformer architectures. The core architecture breaks down transformer components into modular, reusable units that can be easily assembled and modified. This approach enables researchers to experiment with architectural variations while maintaining code readability and understanding of the underlying mechanisms.

The implementation utilizes PyTorch as its foundation and comprises distinct classes for each major transformer component. These components include multi-head attention mechanisms, feed-forward networks, normalization and dropout layers (implemented together as a "sublayer unit"), encoder/decoder layers, and complete encoder/decoder structures [**Figure 1B**]. Additionally, we provide both greedy and beam search generators for sequence generation tasks.

## Key Features

### Flexible Positional Encoding Framework

A key architectural feature is the implementation of a positional embedding strategy pattern that manages various numeric embedding approaches. A strategy manager serves as an intermediary for selecting and applying different positional encoding implementations within the transformer architecture, allowing for seamless integration of different approaches without requiring modifications to the core architecture.

Our implementation currently supports four distinct positional encoding strategies: sinusoidal, learned, rotary, and ALiBi embeddings, chosen as representative examples of popular approaches in the field. Each strategy is implemented as a separate class and managed through the embedding strategy manager, with the architecture designed to readily accommodate additional encoding strategies as they emerge. This flexibility is particularly valuable because different positional encoding strategies require fundamentally different implementations within the transformer architecture: sinusoidal and learned positional embeddings are added to input token vectors, rotary positional embedding requires adjusting the query and key matrices in the attention calculation, and ALiBi adds values to the attention score matrix (the product of the query and key matrices) [**Figure 1C**]. Our extensible design allows users to activate or deactivate these varied encoding strategies independently, enabling direct comparisons of their effectiveness in various applications, while also providing a framework for implementing and testing novel positional encoding approaches.

## Modular Attention Mechanisms

The multi-head attention implementation allows for varying attention methods to be specified at initialization. This design choice facilitates future extensions, such as the incorporation of sparse attention patterns like Longformer [24] or Big Bird [25], while maintaining a consistent interface for the rest of the architecture [**Supplemental Figure 1**].

## Neural Architecture Search Compatibility

The modular design of AttentionSmithy facilitates automated architecture optimization through neural architecture search (NAS). Components can be easily swapped or modified programmatically, allowing for systematic exploration of architectural variations while maintaining code interpretability.

The NAS workflow was based on the "Multi-Objective NAS with Ax" workflow tutorial on the official PyTorch website [26], utilizing Meta's Ax package [27] to do so. The process includes designing a search space as a separate python script that accepts variables that dictate the model structure, setting up a torchx runner and scheduler for submitting model training scripts concurrently, and defining optimization requirement configurations. Ax uses Bayesian optimization to evaluate and compare model configurations and their predictive accuracy, highlighting the impact specific architecture decisions have on the final loss.

As the BLEU score [28] (reported on a scale of 0-100) was used as the primary metric in the original transformer paper [1], we used it as the evaluation metric for the NAS. The search space used for the machine translation task consisted of 6 adjustable parameters. Each of the four implemented positional encoding methods were able to be switched on or off, the dropout rate, and the activation function used in the feed-forward network were adjusted to optimize the BLEU score. Models were trained for 5 epochs during the search to reduce time complexity.

# Code Availability

The source code for AttentionSmithy is publicly available on GitHub (https://github.com/xomicsdatascience/AttentionSmithy). The code implementing machine translation is available at https://github.com/xomicsdatascience/machine-translation and utilizes the WMT14 German-English dataset [29] accessed through the Hugging Face datasets library. The code implementing geneformer [30] is available at https://github.com/xomicsdatascience/geneformer, utilizing preprocessed data from the original geneformer implementation HuggingFace repository. All code repositories are released under the MIT license. The software originated from a re-implementation of code depicted in the Annotated Transformer article [31].

AttentionSmithy is implemented in Python using PyTorch [15]. To enhance usability and standardization, AttentionSmithy is designed to be compatible with PyTorch Lightning [32], allowing researchers to easily incorporate training loops, distributed training, and other advanced features while maintaining clean, research-focused code.

## LLM assistance

Claude 3.5 Sonnet and chatGPT o1 were used to help with writing this manuscript.

# Results

## Validation Studies

We conducted three validation studies to demonstrate the efficacy and versatility of AttentionSmithy: a replication of the original vanilla transformer model, an optimized model determined by a neural architecture search (NAS), and a bioinformatics application.

### Original Transformer Replication

We implemented the transformer architecture as described in "Attention Is All You Need" with two practical constraints: utilization of a single A100 GPU and a maximum context window of 100 tokens. These constraints were imposed due to computational resource limitations. Like the original paper, we trained on the WMT 2014 English-German dataset, which consists of 4.51M sentence pairs (approximately 9.03M sentences total). Due to our context window restriction, we had to truncate 63,227 sentences that exceeded 100 tokens. After 40 epochs of training, our vanilla implementation achieved a BLEU score of approximately 21 on the machine translation task [**Figure 2A,** red line]. While this falls short of the original paper's BLEU score of 25, it represents a reasonable achievement given the restricted context window and computational resources.

To highlight the ease of NAS enabled with AttentionSmithy, we designed a search space around the original model components to identify architectural changes that may enhance performance. The optimized model from NAS had a more rapidly increasing BLEU score across training steps, and in the end achieved a higher BLEU score of approximately 24 after 40 epochs [**Figure 2A**, blue line], approaching the performance of the original implementation despite our hardware constraints. Key modifications identified from NAS included: simultaneous utilization of all four positional encoding strategies (sinusoidal, learned, rotary, and ALiBi), removal of dropout (reduction from 0.1 to 0.0), and replacement of ReLU with tanh activation in feed-forward networks. This led to generally better translations, an example of which is shown in **Figure 2B**.

### Bioinformatics Application

To demonstrate domain adaptability, we replicated the Geneformer model for transfer learning for transcriptomic single-cell data tasks [30]. Following the original paper's methodology in designing the model, we pre-trained it using a BERT-style architecture on rank-based transcript data, with genes serving as tokens. To verify that the model captures contextual information even after plateauing [**Figure 2C**], we fine-tuned it for cell type classification using this dataset, freezing the first two pre-trained layers and adding a classification layer as specified in their methodology. We used their published human_dcm_hcm_nf dataset for this task, which contains 579,159 cells representing 21 distinct cell types from cardiac tissue from 29 individuals. This implementation achieved over 95% accuracy on the validation dataset, demonstrating successful replication of the Geneformer architecture's ability to transfer contextual relationship information for downstream gene expression analyses [**Figure 2D**].

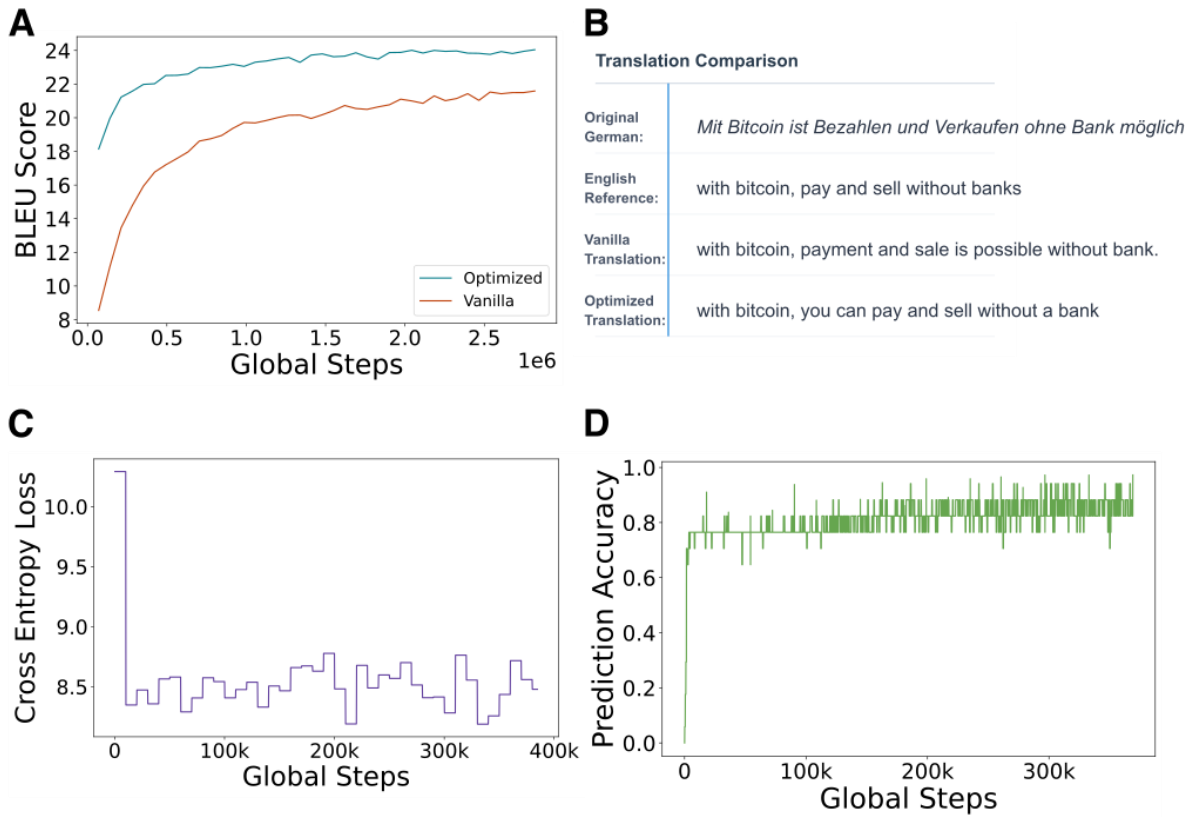

**Figure 2. Validation studies of programs built with AttentionSmithy.** (A) BLEU scores comparing translation quality between a vanilla transformer architecture [1] and an optimized transformer architecture derived through neural architecture search. (B) Representative examples of German-to-English translation outputs, showing source text, reference translation, and outputs from both transformer variants. The BLEU score in this specific example jumped from 16.1 for the vanilla translation to 34.4 in the optimized translation. (C) Validation loss trajectory during pretraining of the Geneformer foundation model plotted against global training steps. (D) Cell type classification accuracy on the validation set during fine-tuning of the pretrained Geneformer model, also plotted against global training steps.

# Discussion

The development and validation of AttentionSmithy reveals several important insights about transformer architecture implementation and customization. Our findings not only demonstrate the software package's effectiveness but also highlight novel approaches to transformer design that merit further investigation.

## Efficacy of Combined Positional Encodings

Perhaps the most intriguing finding from our neural architecture search was the superior performance achieved through the simultaneous application of all four available positional encoding methods. While previous research has typically focused on comparing and contrasting different encoding strategies, our results suggest that these methods may capture complementary aspects of positional information. This discovery opens new avenues for research into how different encoding strategies might interact and complement each other, potentially leading to more robust transformer architectures.

The modular implementation of positional encodings in AttentionSmithy extends beyond traditional position representation. Our framework enables the application of these encoding methods to any numeric data type, offering new possibilities for representing temporal, quantitative, or other ordered information within transformer architectures. For instance, in time-series analysis, one could simultaneously encode both sequential position and temporal information using different encoding strategies, potentially capturing both local and global patterns more effectively.

## Implications for Domain-Specific Applications

While our validation studies focused on established architectures, they serve primarily to demonstrate AttentionSmithy's foundational reliability. The package's true value lies in enabling researchers to develop entirely new transformer architectures for specialized applications that may not yet exist. By providing a flexible, modular framework, we empower domain experts to experiment with novel combinations of transformer components without requiring deep expertise in transformer implementation details.

This capability is particularly valuable in scientific domains where traditional transformer architectures may not perfectly fit the underlying data structures or research questions. For instance, researchers working with complex multimodal data could leverage our framework to develop hybrid architectures that process different data types through specialized attention mechanisms. The ability to experiment with multiple positional encoding strategies simultaneously opens new possibilities for representing complex relationships in data, whether they be spatial, temporal, or domain-specific ordered relationships.

The modular nature of AttentionSmithy enables researchers to focus on the unique aspects of their application domains rather than becoming entangled in transformer implementation details. This democratization of transformer development has the potential to accelerate innovation in fields where artificial intelligence applications are still emerging. For example, researchers could apply self-supervised speech representation techniques [33] to nanopore sequencing, enabling efficient and accurate nucleotide sequencing through pre-training and fine-tuning approaches. In mass spectrometry, developing foundation models to interpret data-independent acquisition (DIA) spectra could allow researchers to leverage these complex, chimeric signals for downstream tasks without relying on pre-existing spectral libraries. The transformative potential of this architecture extends well beyond current applications, and we anticipate that researchers across diverse scientific domains will develop innovative implementations that we cannot yet foresee.

Future development of AttentionSmithy will focus on expanding its capabilities to support emerging transformer variants while maintaining its commitment to architectural clarity and ease of use. We encourage contributions from the research community, particularly in implementing new positional encoding strategies and exploring applications in specialized domains. This could include relative positional embeddings [34] and their T5 variant [35], which focus explicitly on the relationships between positions rather than absolute positions. Through continued development and collaboration, we aim to further lower the barriers to entry for transformer architecture experimentation and innovation across scientific disciplines.

## Acknowledgements

We thank Okan Ozdemir for help understanding HPC resources, and Alexandre Hutton for providing readability comments about the code. This work was funded in part by the NIH (R35GM142502).

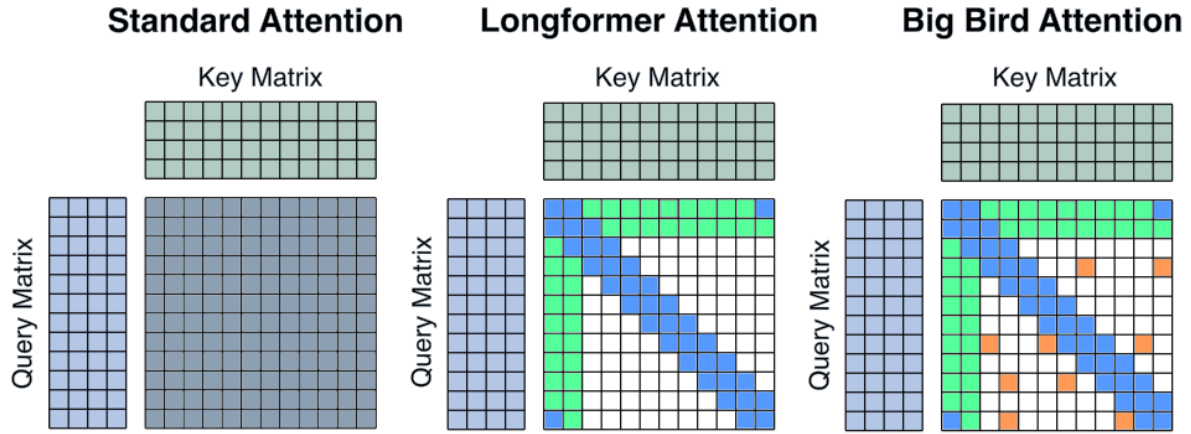

**Supplemental Figure 1: Extendable attention mechanism implementation of AttentionSmithy.** While no alternates are yet implemented, the code is designed to allow alternate attention mechanisms that address various shortcomings of the original method. Two examples for future implementation include the Longformer attention method and Big Bird attention method. Both are designed to extend the allowable context window, a major bottleneck in transformer-based models. This is done by creating a sparse attention matrix only utilizing global and local tokens (both methods) as well as randomly selected tokens (Big Bird only).